# One-Shot Learning with Triplet Loss for Vegetation Classification Tasks[1]

Alexander Uzhinskiy[a], Gennady Ososkov[a], Pavel Goncharov[a],
Andrey Nechaevskiy[a], Artem Smetanin[b]

[a] Joint Institute for Nuclear Research, 6 Joliot-Curie, Dubna, Moscow region, 141980, Russia
[b] ITMO University, Kronverkskiy pr. 49, Saint Petersburg, 197101, Russia

Corresponding author at: Joint Institute for Nuclear Research, 6 Joliot-Curie, Dubna, Moscow region, 141980, Russia. E-mail: auzhinskiy@jinr.ru (A. Uzhinskiy)

## Abstract

Triplet loss function is one of the options that can significantly improve the accuracy of the One-shot Learning tasks. Starting from 2015, many projects use Siamese networks and this kind of loss for face recognition and object classification. In our research, we focused on two tasks related to vegetation. The first one is plant disease detection on 25 classes of five crops (grape, cotton, wheat, cucumbers, and corn). This task is motivated because harvest losses due to diseases is a serious problem for both large farming structures and rural families. The second task is the identification of moss species (5 classes). Mosses are natural bioaccumulators of pollutants; therefore, they are used in environmental monitoring programs. The identification of moss species is an important step in the sample preprocessing. In both tasks, we used self-collected image databases. We tried several deep learning architectures and approaches. Our Siamese network architecture with a triplet loss function and MobileNetV2 as a base network showed the most impressive results in both above-mentioned tasks. The average accuracy for plant disease detection amounted to over 97.8% and 97.6% for moss species classification.

## 1. Introduction

Deep neural networks (DNN) have started a revolution in object detection and recognition. Since 2012, many DNN architectures have been conceived and applied to different tasks including face and object recognition, plant and animal classification, image reconstruction and synthesis, etc. The transfer learning (TL) approach, within which a well-designed DNN model developed for a recognition task is reused as a feature extractor for a new model, is a common way of solving classification tasks. To get good results in a certain domain, researchers must choose an appropriate base network, fine-tune the model, select an optimizer, and a loss function. In our study, we dealt with two vegetation classification tasks. We have created the platform for plant diseases detection (PDD, pdd.jinr.ru) (Uzhinskiy et al., 2019a) in 2017. Crop losses is always a serious problem that costs billion dollars a year to the farmers' community. When we just started our project, there was only one real-life application allowing users to send photos and text descriptions of diseases plants and get a diagnosis and treatment recommendations (Plantix). To obtain the most efficient result we tried various deep learning (DL) architectures and faced the problem of the poor image database (Goncharov et al., 2018). Therefore, we had to collect our own database consisting of 15 classes of three crops, 611 images in total, and to use Siamese networks with cross-entropy loss that allowed us to get 95.7% accuracy (Goncharov et al., 2020). It was not the worthy results, so we planned to extend the database and to look for ways to improve accuracy. Applying the triplet loss function, which showed good results in object classification (Schroff et al., 2015) was one of the options. One more of our projects is focused on the development of the data management system for the ICP Vegetation Program (Uzhinskiy et al., 2019b). This program is carried in the framework of the UNECE Convention on Long-Range Transboundary Air Pollution (LRTAP). The analysis is based on the processing of bio-accumulators (naturally growing mosses) collected in 39 countries of Europe and Asia every 5 years. Contributors responsible for sample collection may not always have the necessary competence to identify the moss species; however, this information is important for the analysis phase. We tried to use the approach of our PDD project to identify moss species. There are 599 images of five moss species in the database and we managed to get only 82% accuracy using the Siamese network with cross-entropy loss and less than 60% with TL based on ResNet50. This is not a very impressive result, probably because the sample used did not allow identifying sufficiently significant properties of mosses, but even an expert finds it difficult to distinguish one type of moss from another. Our research is aimed to test whether our approach with the triplet loss function could improve the situation.

---
[1] Preprint to be published in the Computer Optics journal (ISSN 0134-2452)

## 2. Materials and methods

### 2.1 Related works

There are many studies in which DL models are used to detect and classify plant disease symptoms. We can divide them into two types depending on the database used. The first one uses the PlantVillage database, i.e. – 14 crops and 26 diseases, a total of 54.306 images of diseased and healthy plant leaves collected under controlled conditions (Hughes and Salathé, 2015). The second one collects a dataset in the field or on the internet. Using the PlantVillage dataset with the transfer learning approach allows one to get more than 95% accuracy on a test subset, but the accuracy decreases dramatically on field images. For example, Mohanty et al., 2016 use AlexNet and GoogLeNet on the PlantVillage database to classify 26 diseases and obtain 99.35% accuracy on a test subset. However, poor accuracy of 31.4% is achieved for real-life images. Too et al. use VGG 16, Inception V4, ResNet with 50, 101, and 152 layers, and DenseNets on the PlantVillage database. They try a variety of optimization techniques and training approaches to obtain more than 98% accuracy for all TL models except VGG 16, for which the accuracy is over 82%. The authors have not reported tests on real-life images; however, we can assume that the accuracy could be much worse. Arguments for this assumption can be found in (Ferentinos, 2018), where the author uses AlexNet, AlexNetOWTBn, GoogLeNet, Overfeat, VGG architectures on the second edition of the PlantVillage database with 87,848 images of 58 different classes. The reported validation accuracy exceeds 97%. The authors also test the model on real-life images and present an experiment with real-life images added to the training dataset. The accuracy of the model trained on the PlantVillage database on the field test set is over 33% as in (Mohanty et al., 2016). Accuracy of the model trained on the mixed dataset was over 65% on the filed dataset. Using the PlantVillage database to train the model for a real-life application appointed as a bad idea. Thus, studies with self-collected databases seem to us much more interesting. Fuentes et al. 2017 use a self-collected tomato database of 10 classes, 5000 images in total. The authors consider three types of detectors: Faster Region-based Convolutional Neural Network (Faster R-CNN), Region-based Fully Convolutional Network (R-FCN), and Single Shot Multibox Detector (SSD) combined with a feature extractors such as VGG and ResNet along with the SmoothL1 loss function. The Faster R-CNN with VGG-16 as a feature extractor gives an overall average accuracy of 83% for all classes. Türkoğlu and Hanbay, 2019 use various base networks such as AlexNet, VGG16, VGG19, SqueezeNet, GoogleNet, Inceptionv3, InceptionResNetv2, ResNet50, ResNet101, and classifiers such as K-nearest neighbor (KNN), Support vector machine (SVM), Extreme learning machine (ELM). Their self-collected database comprises 1965 high-resolution images of eight different plant diseases of four crops. The highest-level accuracy, which is obtained with the ResNet50 model and SVM classifier, amounts to 97.86%. The results are very impressive. Unfortunately, there is no link to their database. Selvaraj et al., 2019 focus on banana diseases. They have collected a database with 30,952 images of different parts and entire plants in which 9000 images are photos of leaves. The authors use ResNet50, InceptionV2, and MobileNetV1 architectures. The best result on leaves is over 70% accuracy with ResNet50 and InceptionV2. In general, the accuracy of models trained on field data is lower than that of models trained on PlantVillage data. Saleem et al., 2019 have done a comprehensive review of works related to plant diseases and visualization techniques. It should be noted that there is another class of plant disease studies in which images are taken from aerial vehicles or satellite sensors. It is a highly interesting and promising direction; nevertheless, it is beyond the scope of our current research.

Regarding the classification of the moss spices, we managed to find only one related work using the DL approach (Ise et al. 2018). The training data is prepared by the "chopped picture" method when one high-resolution picture is "chopped" into a number of small pictures. The resulting dataset consists of 93,841 images of three moss species. The authors use the LeNet network with SGD solver on NVIDIA DIGITS 4.0. The accuracy for each moss species is 99%, 95%, and 74%. The authors suppose that some moss species are relatively large and have a relatively distinctive well-defined shape, while others are highly amorphous.

We started our work with plant disease detection architectures in 2017. Firstly, we reproduced the TL approach on the PlantVillage database considering only grape diseases. We tried four base networks such as VGG19, InceptionV3, ResNet50, and Xception. ResNet50 showed the best result, i.e. - 99.4%. The accuracy on the test subset of 30 images from the Internet was about 48%. We unsuccessfully tried different ways of data modification described in (Goncharov et al., 2018). We realized that the PlantVillage database was not an option in case of the real-life application. PlantVillage images have the same background, illumination, position, and orientation. In the real-life, images taken in the different conditions under various angles and at the background can be anything. Thus, we began collecting our own database. We started from four classes of grape diseases, a total of 279 images including healthy leaves. We faced a problem of training models on a small amount of data. The solution was found with the help of a One-shot learning approach, namely Siamese network. Being already trained one of the Siamese network twins acts as a feature extractor followed by the KNN-classifier. With this approach, we achieved the accuracy of 94%. After the extension of the database to 15 classes of three crops, 611 images in total, we obtain a lower accuracy of 86% while preserving the same classification scheme. We decided that the problem was in the classifier and tried different options. With the Siamese network as a feature extractor and multi-layer perceptron (MLP) as a classifier, we managed to get 95.7% accuracy (Goncharov et al., 2020). The moss classification task appeared in 2019. At that time, we already had some practices. Thus, we started from the TL approach with ResNet50. The accuracy was near 59%. The Siamese network with MLP gave only 78% accuracy. We made some changes to the convolutional layers but could not obtain more than 81% accuracy. We believe that Siamese networks is a very promising direction, and we hoped that we could improve accuracy using the triplet loss function.

**2.2 The triplet loss function**

In (Goncharov et al., 2020) we used Siamese networks with cross-entropy loss to get 95.7% accuracy (Goncharov et al., 2020) on 15 classes. After increasing the database to 25 classes, the accuracy decreased to 89.6%. We wanted to check if the triplet loss could improve accuracy.

Siamese networks and the triplet loss function can be used when there is a paucity of training data available. This combination has shown impressive results in facial recognition tasks (Schroff et al.. 2015; Cheng et al.2016; Hermans et al., 2017), object tracking (Dong and Shen, 2017), brain imaging modality recognition (Puch et al., 2019), bioacoustics classification (Anshul et al., 2019), remote sensing scene classification (Zhang et al., 2020) and other tasks.

The Siamese network consists of a twin network with tied weights joined by the similarity layer with the energy function at the top. When we pass an image to the network input, we extract some features of the image in the output, so-called embeddings. Similar images cannot be in very different locations of the feature space, because each of the twins computes the same function due to weights sharing. The triplet loss function use three images during evaluations. The anchor is an arbitrary data point. The positive image belongs to the same class as the anchor. The negative image belongs to a different class from the anchor. The triple loss reduces the distance between the anchor and the positive image while increasing the distance between the anchor and the negative image (see Fig. 1).

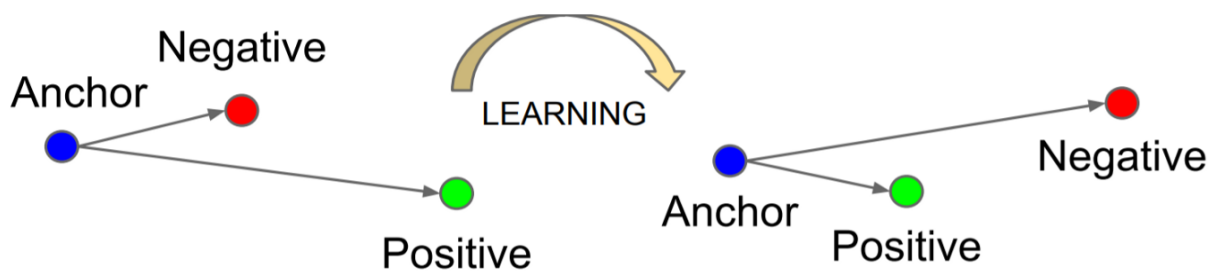

Figure 1. Visualization of the learning process with the triplet loss function

When training the Siamese network with the triplet loss function, the input consists of three images, two of which belong to the same class, and the last one belongs to a different class. The model processes each image and produce a feature vector. In the end, we can say that the two images are from different classes if the distance between them is high and that they are from the same class if the distance is small.

**2.3 Image database**

Our image dataset for plant disease detection was collected from open sources on the Internet. It consists of 25 classes of four crops (cotton, wheat, corn, cucumbers, and grape), 935 images in total (see Fig.2). All images are 256x256 pixels and contain meaningful information about crops and diseases on them. Comparing with our old publication (Goncharov et al., 2020), we have added two new crops, i.e. – cotton and cucumbers with ten classes (Alternaria leaf blight,

Healthy, Powdery mildew, Verticillium wilt, Nutrient deficiency for cotton and Anthracnose, Downy mildew, Healthy, Nutrient deficiency, Powdery mildew for cucumbers). The database and the link for downloading can be found at http://pdd.jinr.ru/.

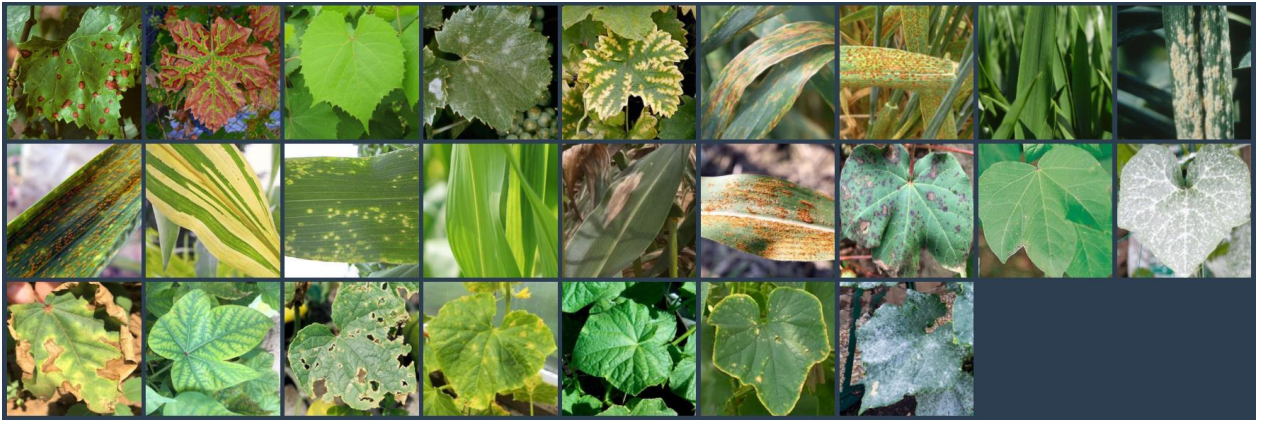

Figure 2. Images of 20 classes of diseases from the PDD database.

Our moss species dataset consists of 599 images of five species (Abietinella Abietina, Hylocomium Splendens, Hypnum Cupressiforme, Pleurozium Schreberi, and Pseudoscleropodium Purum), see Fig.3. All images are 256x256 pixels. The database and the link for downloading can be found at http://moss.jinr.ru/.

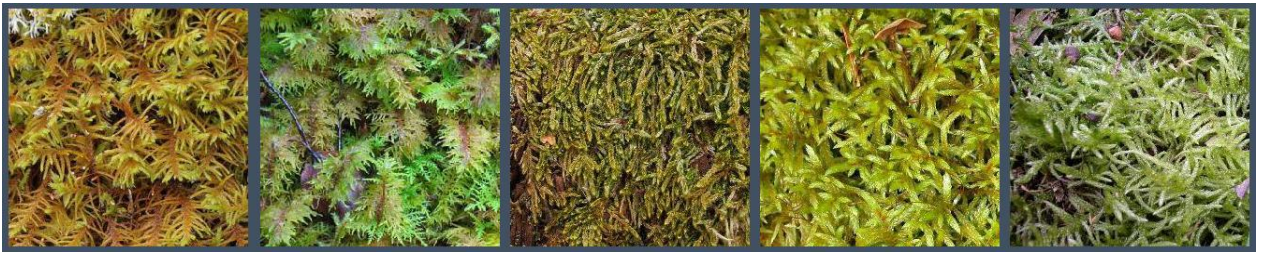

Fig. 3. Images of five moss spices.

**2.4 Current solution**

In our current architecture, we have a Siamese network with three twins. We use MobileNetV2 pre-trained on the ImageNet dataset as a base network for twins. The Siamese network is trained with the help of triplet loss using the TripletTorch utility. After training, one of the twins is used as a feature extractor for a multi-layer perceptron, which acts as a classifier (see Fig. 4).

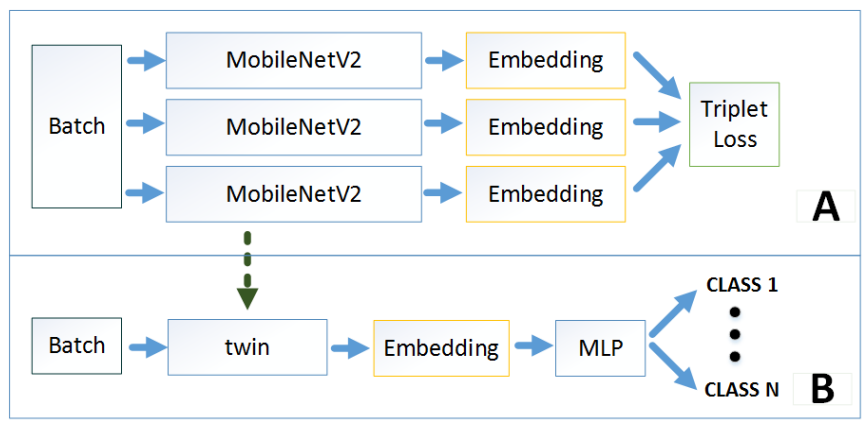

Fig. 4. Current architecture. A – Siamese network with three twins and the triplet loss function. B - one of the twins used as a feature extractor and MLP used as a classifier.

The model was trained on the "HybriLIT" heterogeneous platform of the Joint Institute for Nuclear Research. The platform has various computing architectures, and we used NVIDIA Tesla K40. We trained the Siamese network for 300 epochs with a batch size 32 and a learning rate of 0.001. We used an 80 to 20 ratio to split the data into train and test subset. It took about 40 minutes to train the model. The Adam optimizer with learning rate equals to 0.001 and cross-entropy loss was used to train the perceptron. The classifier model was trained for 40 epochs, and it took a couple of minutes.

At present, each model weights over 18 Mb. We want to have a model that can be run "on board" of mobile devices; therefore, we have tried dynamic and static quantization with eager mode. Formally, we have trained the feature extractor, quantized it, and train one more epoch for weights calibration. Then the quantized model is used to train the classifier. We manage to reduce the model size to over 7 Mb and increase the performance by 5 times with static quantization. Without quantization consecutive processing of 100 images take approximately 14.2 seconds. The quantized model consecutively processes 100 images in less than 2.6 seconds.

**3. Results and discussion**

We trained the model for plant disease detection 16 times each time with a new split into train and validation. The average accuracy was 97.8%, while the highest accuracy was 99.4%. The moss classification model in the same condition showed 97.65% average accuracy and 100% highest accuracy.

With TL based on ResNet50, we could not obtain accuracy more than 60% for both tasks. The accuracy of siamese networks with cross-entropy loss on plant disease detection task was over 89% and less than 81% for the moss classification task. With the triplet loss function, we can get the state of the art results for both tasks. Fig. 5 shows the representation of the extracted feature vectors in 2D space for the moss spices model before training and after 300 epochs of training. For the plant disease detection model the situation is similar but many classes are blocked off; however, we can see a clear separation on the others.

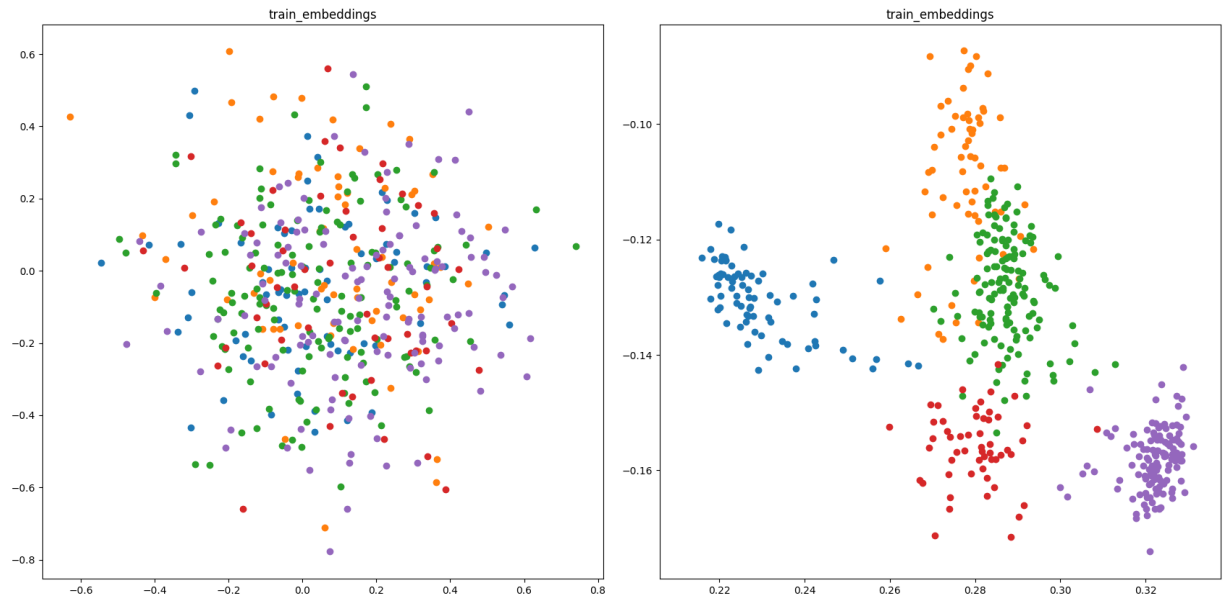

Fig. 5. Representation of the extracted feature vectors of 5 moss classes in 2D. A – before training. B – after 300 epochs of training.

During the evaluation of (Uzhinskiy et al., 2019a) we collected a test dataset of images to compare our model with the well-known commercial AutoML-solution. The dataset had 60 images consisting of 15 classes. For each class, we had two images that were used for training and two new for the model images. We expanded the dataset with 40 images of five new classes of cotton diseases. Our old model (Siamese networks with cross-entropy loss and MLP classifier) achieved 88.3% accuracy on the test dataset with 15 classes. The new model showed 98% accuracy on the test dataset with 25 classes. We are focused on the development of production-ready solution for plant disease detection and we have some tests for our model. One of the tests is the evaluation of the model on all our images. Fig. 6 shows the confusion matrix for our current production model. There are 30 wrongly classified images of 935 images in total. One can see that most of the wrong recognition is for the same disease on the different crops or the diseases that look similar.

| | Cotton__Alternaria leaf blight | Cotton__Healthy | Cotton__Nutrient deficiency | Cotton__Powdery mildew | Cotton__Verticillium wilt | Grape__Black rot | Grape__Chlorosis | Grape__Esca | Grape__Healthy | Grape__Powdery mildew | Corn__Downy mildew | Corn__Eyespot | Corn__Healthy | Corn__Northern leaf blight | Corn__Southern rust | Wheat__Black chaff | Wheat__Brown rust | Wheat__Healthy | Wheat__Powdery mildew | Wheat__Yellow rust | Cucumbers__Anthracnose | Cucumbers__Downy mildew | Cucumbers__Healthy | Cucumbers__Nutrient deficiency | Cucumbers__Powdery mildew |
|---|---|---|---|---|---|---|---|---|---|---|---|---|---|---|---|---|---|---|---|---|---|---|---|---|---|
| Cotton__Alternaria leaf blight | 33 | 0 | 0 | 1 | 0 | 0 | 0 | 0 | 0 | 0 | 0 | 0 | 0 | 0 | 0 | 0 | 0 | 0 | 0 | 0 | 0 | 0 | 0 | 0 | 0 |
| Cotton__Healthy | 0 | 31 | 0 | 0 | 0 | 0 | 0 | 0 | 0 | 0 | 0 | 0 | 0 | 0 | 0 | 0 | 0 | 0 | 0 | 0 | 0 | 0 | 0 | 0 | 0 |
| Cotton__Nutrient deficiency | 0 | 0 | 31 | 0 | 0 | 0 | 0 | 0 | 0 | 0 | 0 | 0 | 0 | 0 | 0 | 0 | 0 | 0 | 0 | 0 | 0 | 0 | 0 | 0 | 0 |
| Cotton__Powdery mildew | 1 | 0 | 0 | 28 | 0 | 0 | 0 | 0 | 0 | 0 | 0 | 0 | 0 | 0 | 0 | 0 | 0 | 0 | 0 | 0 | 0 | 0 | 0 | 0 | 1 |
| Cotton__Verticillium wilt | 0 | 0 | 0 | 0 | 30 | 0 | 0 | 0 | 0 | 0 | 0 | 0 | 0 | 0 | 0 | 0 | 0 | 0 | 0 | 0 | 0 | 0 | 0 | 0 | 0 |
| Grape__Black rot | 0 | 0 | 0 | 0 | 0 | 31 | 0 | 0 | 0 | 0 | 0 | 0 | 0 | 0 | 0 | 0 | 0 | 0 | 0 | 0 | 0 | 0 | 0 | 0 | 0 |
| Grape__Chlorosis | 0 | 0 | 0 | 0 | 0 | 0 | 49 | 0 | 0 | 0 | 0 | 0 | 0 | 0 | 0 | 0 | 0 | 0 | 0 | 0 | 0 | 0 | 0 | 0 | 0 |
| Grape__Esca | 0 | 0 | 0 | 0 | 0 | 0 | 0 | 73 | 0 | 0 | 0 | 0 | 0 | 0 | 0 | 0 | 0 | 0 | 0 | 0 | 0 | 0 | 0 | 0 | 0 |
| Grape__Healthy | 0 | 0 | 0 | 0 | 0 | 0 | 0 | 0 | 121 | 0 | 0 | 0 | 0 | 0 | 0 | 0 | 0 | 0 | 0 | 0 | 0 | 0 | 0 | 0 | 0 |
| Grape__Powdery mildew | 0 | 0 | 0 | 0 | 0 | 0 | 0 | 0 | 0 | 22 | 0 | 0 | 0 | 0 | 0 | 0 | 0 | 0 | 0 | 0 | 0 | 0 | 0 | 0 | 1 |
| Corn__Downy mildew | 0 | 0 | 0 | 0 | 0 | 0 | 0 | 0 | 0 | 0 | 33 | 0 | 0 | 0 | 0 | 0 | 1 | 0 | 0 | 0 | 0 | 0 | 0 | 0 | 0 |
| Corn__Eyespot | 0 | 0 | 0 | 0 | 0 | 0 | 0 | 0 | 0 | 0 | 0 | 31 | 0 | 0 | 0 | 0 | 0 | 0 | 0 | 0 | 0 | 0 | 0 | 0 | 0 |
| Corn__Healthy | 0 | 0 | 0 | 0 | 0 | 0 | 0 | 0 | 0 | 0 | 0 | 0 | 34 | 0 | 0 | 0 | 0 | 0 | 0 | 0 | 0 | 0 | 0 | 0 | 0 |
| Corn__Northern leaf blight | 0 | 0 | 0 | 0 | 0 | 0 | 0 | 0 | 0 | 0 | 0 | 0 | 0 | 34 | 0 | 0 | 0 | 3 | 0 | 0 | 0 | 0 | 0 | 0 | 0 |
| Corn__Southern rust | 0 | 0 | 0 | 0 | 0 | 0 | 0 | 0 | 0 | 0 | 0 | 0 | 0 | 0 | 32 | 0 | 0 | 0 | 0 | 1 | 0 | 0 | 0 | 0 | 0 |
| Wheat__Black chaff | 0 | 0 | 0 | 0 | 0 | 0 | 0 | 0 | 0 | 0 | 0 | 0 | 0 | 0 | 0 | 29 | 1 | 0 | 0 | 0 | 0 | 0 | 0 | 0 | 0 |
| Wheat__Brown rust | 0 | 0 | 0 | 0 | 0 | 0 | 0 | 0 | 0 | 0 | 0 | 0 | 0 | 0 | 0 | 1 | 30 | 0 | 0 | 1 | 0 | 0 | 0 | 0 | 0 |
| Wheat__Healthy | 0 | 0 | 0 | 0 | 0 | 0 | 0 | 0 | 0 | 0 | 1 | 0 | 0 | 3 | 0 | 0 | 0 | 25 | 0 | 1 | 0 | 0 | 0 | 0 | 0 |
| Wheat__Powdery mildew | 0 | 0 | 0 | 0 | 0 | 0 | 0 | 0 | 0 | 0 | 0 | 0 | 0 | 0 | 0 | 0 | 0 | 0 | 30 | 0 | 0 | 0 | 0 | 0 | 0 |
| Wheat__Yellow rust | 0 | 0 | 0 | 0 | 0 | 0 | 0 | 0 | 0 | 0 | 0 | 0 | 0 | 0 | 1 | 0 | 1 | 1 | 0 | 29 | 0 | 0 | 0 | 0 | 0 |
| Cucumbers__Anthracnose | 0 | 0 | 0 | 0 | 0 | 0 | 0 | 0 | 0 | 0 | 0 | 0 | 0 | 0 | 0 | 0 | 0 | 0 | 0 | 0 | 30 | 2 | 1 | 0 | 0 |
| Cucumbers__Downy mildew | 0 | 0 | 0 | 0 | 0 | 0 | 0 | 0 | 0 | 0 | 0 | 0 | 0 | 0 | 0 | 0 | 0 | 0 | 0 | 0 | 2 | 33 | 0 | 1 | 2 |
| Cucumbers__Healthy | 0 | 0 | 0 | 0 | 0 | 0 | 0 | 0 | 0 | 0 | 0 | 0 | 0 | 0 | 0 | 0 | 0 | 0 | 0 | 0 | 1 | 0 | 36 | 0 | 0 |
| Cucumbers__Nutrient deficiency | 0 | 0 | 0 | 0 | 0 | 0 | 0 | 0 | 0 | 0 | 0 | 0 | 0 | 0 | 0 | 0 | 0 | 0 | 0 | 0 | 0 | 1 | 0 | 32 | 0 |
| Cucumbers__Powdery mildew | 0 | 0 | 0 | 1 | 0 | 0 | 0 | 0 | 0 | 1 | 0 | 0 | 0 | 0 | 0 | 0 | 0 | 0 | 0 | 0 | 0 | 2 | 0 | 0 | 30 |

Figure 6. Confusion matrix for 935 images from plant disease dataset.

Crop losses due to diseases is a serious problem; however, the number of real-life applications allowing to detect the diseases is very limited. Owing to the lack of production field image datasets, it is necessary to improve One-shot and Few-shot learning methods. Our work shows that even with a limited dataset it is possible to train a good model. We have already deployed the new model into our PDD portals and mobile application, thus it is available to the farmers' community.

Air pollution has a significant negative impact on various components of ecosystems and human health. Moss as a bioaccumulators is a great data source for environmental monitoring. The identification of moss species is important for the quality of the analysis. We have deployed our new model into an application that allows filling in the information about sampling sites required by the UNECE ICP Vegetation manual. We hope this will help the contributors of the 2020-2022 survey to classify moss species right

The accuracy of the resulting model shows that the combination of Siamese networks with the triplet loss function is highly promising in the case of vegetation classification tasks. We have shown effectiveness on two tasks; however, we believe that such an approach can be used in many other applications.

For us now, it is impossible to create a dataset for detector algorithms like Fast R-CNN, Faster R-CNN, YOLO, but we would like to have an ability to recognize disease on a part of an image. The processing time of our model decreased after quantization, thus, we are going to build an R-CNN object detector based on our model and OpenCV abilities (selective search, confidence filtering, and non-maxima suppression). We will continue to expand our databases and improve our models; event though, we already have good results for both of our tasks.

## 4. Conclusion

We applied the Siamese networks along with the triplet loss function to solve two vegetation classification tasks. The model for plant disease detection trained on 25 classes of five crops shows 97.8% accuracy. The model for moss species classification on five classes shows 97.6% accuracy. This result is much better than the transfer learning approach itself with a base network like ResNet50 and better than our previous approach with the Siamese network with two twins and with cross-entropy loss. The combination of the Siamese network with the triplet loss function has a great potential for classification tasks with a very small training dataset. We had only 935 images of diseased plants and 599 images of mosses; nevertheless, we manage to obtain good results. The dataset and the models are accessible via pdd.jinr.ru and moss.jinr.ru. We are going to expand our databases, improve our models, and build an R-CNN object detector based on our model and OpenCV.

## 5. Acknowledgements

The reported study was funded by RFBR according to the research project № 18-07-00829.